# Hybrid Facial Expression Recognition (FER2013) Model for Real-Time Emotion Classification and Prediction

Ozioma Collins Oguine[1, *], Kanyifeechukwu Jane Oguine[2], Hashim Ibrahim Bisallah[3], Daniel Ofuani[4]

[1]Department of Computer Science, University of Abuja, Abuja, Nigeria

**Email address:**
oziomaoguine007@gmail.com (O. C. Oguine)

**Abstract:** Facial Expression Recognition is a vital research topic in most fields ranging from artificial intelligence and gaming to Human-Computer Interaction (HCI) and Psychology. This paper proposes a hybrid model for Facial Expression recognition, which comprises a Deep Convolutional Neural Network (DCNN) and Haar Cascade deep learning architectures. The objective is to classify real-time and digital facial images into one of the seven facial emotion categories considered. The DCNN employed in this research has more convolutional layers, ReLU Activation functions, and multiple kernels to enhance filtering depth and facial feature extraction. In addition, a haar cascade model was also mutually used to detect facial features in real-time images and video frames. Grayscale images from the Kaggle repository (FER-2013) and then exploited Graphics Processing Unit (GPU) computation to expedite the training and validation process. Pre-processing and data augmentation techniques are applied to improve training efficiency and classification performance. The experimental results show a significantly improved classification performance compared to state-of-the-art (SoTA) experiments and research. Also, compared to other conventional models, this paper validates that the proposed architecture is superior in classification performance with an improvement of up to 6%, totaling up to 70% accuracy, and with less execution time of 2098.8s.

## 1. Introduction

Humans possess a natural ability to understand facial expressions. In real life, humans express the emotions on their faces to show their psychological state and disposition at a time and during their interactions with other people. However, the current trend of transferring cognitive intelligence to machines has stirred up conversations and research in the domain of Human-Computer Interaction (HCI) and Computer Vision, with a particular interest in Facial Emotion Recognition and its application in human-computer collaboration, data-driven animation, human-robot communication, etc.

Emotions are physical and instinctive, instantly prompting bodily reactions to threats, rewards, and other environmental factors. Responses to these factors are primarily based on objective measurements such as pupil dilation (eye-tracking), skin conductance (EDA/GSR), voice analysis, body language analysis, gait analysis, brain activity (fMRI), heart rate (ECG), and facial expressions. A notable ability for humans to interpret emotions is crucial to effective communication; hypothetically, 93% of efficient conversation depends on the emotion of an entity. Hence, for ideal Human-computer interaction (HCI), a high-level understanding of the human emotion is required by machines.

Emotions are a fundamental part of human communication, driven by the erratic nature of the human mind and the perception of relayed information from the environment. There are varied emotions that inform decision-making and are vital components in individual reactions and psychological state. Contemporary psychological research observed that facial expressions are predominantly used to understand social interactions rather than the psychological state or personal emotions. Consequently, the credibility assessment of facial expressions, which includes the discernment of genuine (natural) expressions from postured (deliberate/volitional/deceptive) expressions, is a crucial yet challenging task in facial emotion recognition. This research will focus on educating objective facial parameters from real-time and digital images to determine the emotional states of people given their facial expressions, as shown in Figure 1.

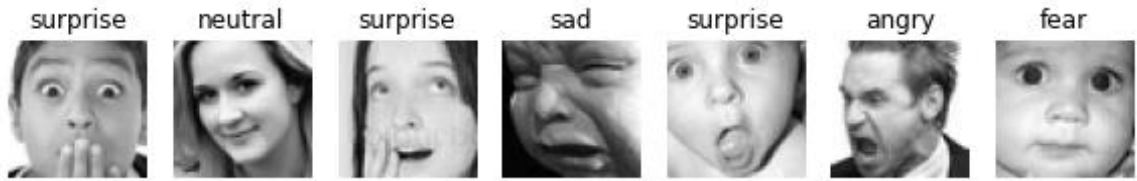

**Figure 1.** *FER-2013 Sample Training Set Images.*

Over the last six years, advancement in deep learning has propelled the evolution of convolutional networks. Computer Vision is an interdisciplinary scientific field that equips the computer with a high-level understanding from images or videos that replicate human visual prowess within a computer perspective. It aims to automate tasks, categorize images from a given set of classes, and have the network determine the predominant class present in the image. It can be implied that computer vision is the art of making a computer 'see' and impacting it with human intelligence of processing what is 'seen'. More recently, deep learning has become the go-to method for image recognition and detection, far usurping medieval computer vision methods due to the unceasing improvement in the state-of-the-art performance of the models.

A gap identified by this research is with regards to the fact that most datasets of preceding research consist of well-labeled (posed) images obtained from a controlled environment, usually postured. According to Montesinos López et al., this anomaly increased the likelihood of model overfitting, given insufficient training data availability, ultimately causing a relatively lesser efficiency in predicting emotions in uncontrolled scenarios [1]. Consequently, this research also identified the importance of lighting in facial emotion recognition (FER), highlighting that poor lighting conditions could decline the model's efficiency.

This research will use Convolutional Neural Networking (CNN) to model some critical extracted features used for facial detection and classify the human psychological state into one of the six emotions or a seventh neutral emotion. It also will employ the Haar Cascade model for real-time facial detection. It is worthy of note that given the hand-engineered nature of the features and model dependency on prior knowledge, a resultant comparatively higher model accuracy is required.

## 2. Empirical Review of Related Work

Over the years, several scholars have embarked on research to tackle this novel challenge. Ekman & Friesen, highlighted seven basic human emotions independent of the culture a human being is born into (anger, fear, disgust, happiness, sad, surprise, and neutral) [2]. Sajid et al., in a recent study on the Facial Recognition Technology (FERET) dataset, discovered the influence of facial asymmetry as a marker for age estimation, which emphasized the ease in detecting the right face asymmetry when compared to the left face [3]. Below are some reviews of works of literature pertinent to this research.

Kahou et al. used a CNN-RNN combined architecture to train a model on individual frames of videos and static digital images [4]. The Acted Facial Expressions in the Wild (AFEW) 5.0 dataset was utilized for the video clips and a combination of the FER-2013 and Toronto Face Database for the images [5]. As an alternative to using long short-term memory (LSTM) units, IRNNs were used, which consists of rectified linear units (ReLUs) [6]. IRNNs were suitable because they provided a simple mechanism for dealing with the vanishing and exploding gradient problem. The research achieved an overall accuracy of 0.528.

Facial pose appearance is still a critical problem in face detection. Ratyal et al., provided the solution for variability in facial pose appearance. Using subject-specific descriptors, three-dimensional pose invariant approach was used [7, 8].

Ming Li et al. propose a neural network model to overcome two shortcomings in still image-based FERs: the inter-variability of emotions between subjects and misclassification [9]. Their model consists of two convolutional neural networks - the first is trained with facial expression databases, while the second is a DeepID network used to learn identity features. These two neural networks were then concatenated together as a Tandem Facial Expression of TFE Feature, passed to the fully connected layers to form a new model. In Mou et al., the authors inferred group emotion using arousal and valence emotion annotations using face, body, and context features [10]. Tan et al. used a combination of two types of CNNs, namely individual facial emotion CNNs and global image-based CNNs, to recognize group-level emotion in images [11]. Different pooling methods such as average and max pooling are used to downsample the inputs and aid in generalization [12, 13]. Dropout, regularization, and data augmentation were used to prevent overfitting. Batch normalization was developed to help prevent gradient vanishing and exploding [14, 15].

As can be inferred from the literature highlighted above, several innovative research conducted by other scholars has emphasized continuous upscaling of accuracy without consideration for efficiency simultaneously. This research paper proposes a more efficient and accurate model with improved generalization, as discussed in subsequent sections. Table 1 summarizes previous reported classification accuracies FER2013. Most reported methods perform better than the estimated human performance (~ 65.5 %). In this work, a state-of-the-art accuracy of 70.04 % was achieved.

**Table 1.** *Summary of previous reported Accuracies for the FER-2013 Dataset.*

| Methods | Accuracy Rating |
|---|---|
| CNN [16] | 62.44% |
| GoogleNet [19] | 65.20% |
| VGG + SVM [18] | 66.31% |
| Conv + Inception Layer [20] | 66.40% |
| Bag of words [17] | 67.40% |
| CNN + Haar Cascade (This work) | 70.04% |

## 3. Theoretical Background

The FER problem categorizes the emotions into classes such as sadness, surprise, anger, happiness, fear, disgust, and neutrality [21]. It constructs a classification algorithm by using extracted features. Researchers in the past have utilized classifiers such as Support Vector Machine (SVM), Artificial Neural Network (ANN), Hidden Markov Model (HMM), and K-Nearest Neighbor (KNN) were used for recognizing facial expressions. This research employs a modified hybrid model, which comprises two models (CNN and Haar Cascade). Several model components are explained below and specify an architecture to learn multiple levels of abstraction and representations. These components include Convolutional layers, dropout, ReLU Activation Function, Categorical Cross-Entropy Loss, Adam Optimizer, and SoftMax Activation Function in the output layer for seven emotion classifications, as shown in Figure 2. This section will also emphasize the modifications made to these components through hyperparameter tuning that enhanced the proficiency and accuracy of the hybrid model.

Convolutional Layer: The Conv layer is the core building block of a Convolutional Neural Network that does the most computational heavy lifting. For a given input, the convolution layer performs a convolution [21] using a filter $fk$ of kernel size $n*m$ and applied to an input x. The number of input connections is $n*m$. The output can be calculated as equation (1).

$$C(X_{U,V}) = \sum_{i=n/2}^{n/2} \sum_{j=-n/2}^{n/2} f_k(i,j) x_{(u-i,v-j)} \quad (1)$$

Max Pooling Layer: It reduces the input by a considerable amount by applying the max function Saravanan et al. [22]. Let $m$ be the size of the filter and $x_i$ be the input. An implementation of this layer is seen in Figure 2, and the output can be calculated as follows in equation (2).

$$M(x_i) = max\ (r_{i+k,i+l}\ /|k| \le m\ /2, |l| < m/2, k, l\ \in N) \quad (2)$$

Rectified Linear Unit (ReLU) Activation Function: It calculates the output for a given value of p, the input to the network or a neuron, as represented in equation (3). ReLU was utilized for this research because it has no exponential function to calculate and has no vanishing gradient error. Figure 2 below shows the concatenated convolutional layers parallelly processed via ReLU activation functions to upscale accuracy and obtain facial features of images flawlessly

$$f(x) = \begin{cases} 0, for\ x < 0 \\ 1, for\ x \ge 0 \end{cases} \quad (3)$$

Fully Connected Layer: This is popularly known multilayer perceptron, and it converts all the neurons of previous layers to each neuron of its layer. It is mathematically represented as the equation (4).

$$f(x) = \sigma(p*x) \quad (4)$$

Where $\sigma$ is the activation function, and $p$ is the resultant matrix of size $n*k$. $k$ is the dimension of $x$, and $n$ is the number of neurons in a fully connected layer.

Output Layer: It represents the class of the given input image and is one hot vector [21]. It is expressed in equation (5).

$$C(x) = (i|\exists v_{j \ne i}\ and\ x_j \le\ x_i) \quad (5)$$

SoftMax layer: Its main functionality is backpropagation of error [21]. Let the dimension of an input vector be P. Then, it can be represented as a mapping function as expressed in equation (6).

$$S(x): R^P \to [0,\quad 1]^P \quad (6)$$

and the output for each component j (1 ≤ j ≤ P) is given as equation (7).

$$S(x)_j = \frac{e^{x_j}}{\sum_{i=1}^{p} e^{x_i}} \quad (7)$$

## 4. Proposed Methodology

With the need for real-time object detection, several object detection architectures have gained wide adoption by most researchers. However, the Hybrid Architecture put forward by this research uses both the Haar Cascade Face Detection, a popular facial detection algorithm proposed by Paul Viola and Michael Jones in 2001, and the CNN Model together. In Figure 2 below, the CNN architecture initially needs to extract input pictures of 48x48x1 (48 wide, 48 high, 1 color channel) from dataset FER-2013. The network starts with an input layer equal to the input data dimension of 48 by 48. It also consists of seven concatenated convolutional layers parallelly processed via ReLU activation functions to upscale accuracy and obtain facial features of images flawlessly, as shown in Figure 2. Sub-models for extracting features share this input and have the same kernel size. The outputs from these feature extraction sub-models are flattened into vectors, concatenated into one lengthy vector matrix, and transmitted to a fully connected layer for evaluation before a final output layer permits classification. A detailed step of the methodology is described in the architecture below.

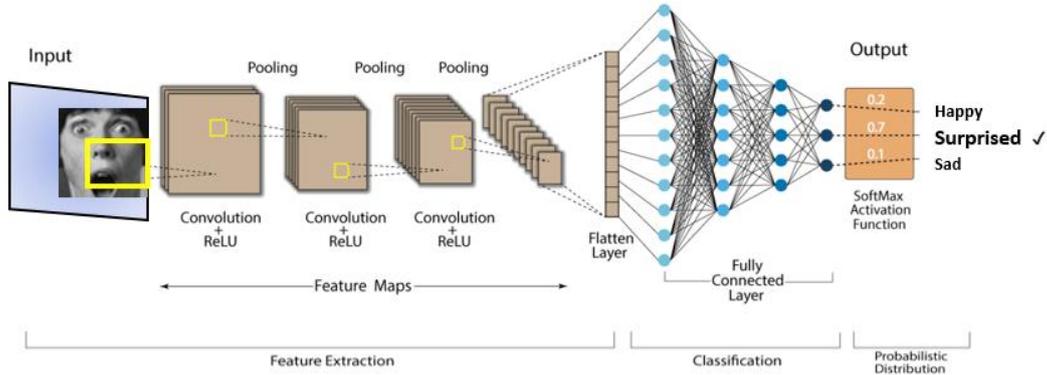

*Figure 2. Proposed CNN Model Structure.*

As seen in Figure 2 above, the proposed CNN model consists of a batch normalization layer followed by 7 convolutional layers with independent learnable filters (kernels), each with sizes of [3x3], and a local contrast normalization layer to remove the average from the neighborhood pixels. It also consists of a max-pooling layer to reduce the spatial dimension of the image, ensuring an increased processing pace and flatten and dense layers. It is followed by a completely connected layer and SoftMax for classifying seven emotions. A dropout of 0.5 for decreasing over-fitting was applied to the fully connected layer, and all layers encompasses rectified linear units (ReLU) activation function. After that, concatenation of two comparable models is linked to a SoftMax output layer that can classify the seven target emotions.

*Real-Time Classification*

The output of the DCNN model is saved as a JSON string file. A JSON or JavaScript Object Notation was deemed suitable for this research because it stores and allows faster data exchange, according to Ingale et al. [23]. The model.tojson() function of python is used to write the output of the trained model into JSON. A Pre-trained haar cascade XML file for frontal face detection of real-time facial feature classification was imported. A multiscale detection approach was implemented, and parameters such as coordinates of bounding boxes around detected faces and detectMultiScale (grayscale inputs) and scale factor were tuned for better efficiency.

In this research, OpenCV's Haar cascade detects and extracts the face region from the webcam's video feed through the flask app. This process follows a video conversion to grayscale, and the detected face is contoured or enclosed within a region surrounding the face (See Figure 3).

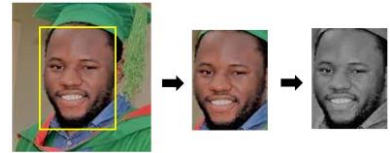

*Figure 3. Facial Detection and Conversion to Grayscale.*

## 5. Experimental Details

*Dataset:*

This research uses the Facial Emotion Recognition (FER-2013) dataset housed by the Kaggle repository. The FER-2013 dataset contains 35887 images, of which 28709 labeled images comprise the training set, while the remaining 7178 images belong to the test set. The images in the FER-2013 dataset are categorized into seven universal emotions: Happy, Sad, Angry, Fear, Surprise, Disgust, and Neutral. The images are in grayscale with 48x48 pixels dimensions. The dataset is summarized in Table 2. A Nvidia Tesla K80 GPU on Google Cloud (Colab Research) is used for training the models. It is a collaborative version of an iPython notebook or Jupyter Notebook.

*Table 2. Summary of the FER-2013 dataset.*

|  | Surprise | Fear | Angry | Neutral | Sad | Disgust | Happy | Total |
|---|---|---|---|---|---|---|---|---|
| train_count | 3171 | 4097 | 3995 | 4965 | 4830 | 436 | 7215 | 28709 |
| Total Count of the test dataset | | | | | | | | |
|  | Surprise | Fear | Angry | Neutral | Sad | Disgust | Happy | |
| test_count | 831 | 1024 | 958 | 1233 | 1247 | 111 | 1774 | 7178 |
|  |  |  |  |  |  |  |  | 35887 |

*Pre-processing*

Face detection and registration operations were applied once to each image sample. They are necessary for this process to correct any existing variations in pose and illumination from the real-time facial detection operation. Keras library was used in this research for the preprocessing of both the test and train images before passing them to the Deep Convolution Neural Network (DCNN). This process includes cropping of detected faces and scaling of the detected images. For Data cleaning purposes, all image target sizes were resized to a dimension of 48x48x1, converted into a grayscale color channel, and preprocessed in a batch size of 120. Square boxes were also used for facial landmark detection to apply illumination correction.

*Training & Validation*

This research used the Keras library in python to accept images to train. To minimize the losses of the Neural Network during training, this research used a Mini-Batch Gradient Descent algorithm. This type of Gradient Descent algorithm was suitable because of its capabilities in finding the coefficients or weights of neural networks by splitting the training dataset into small batches, i.e., a training batch & a validation batch, using Data Augmentation techniques. From Keras, a sequential model was implemented to define the flow of the DCNN, after which several other layers, as described in the proposed methodology above and shown in Figure 2. With an increase in the convolutional layers, a concurrent increase in the kernel size was also adopted, which was used in parallel with the ReLU Activation function for training the model. Dropouts were used at several layers of the DCNN to prevent the overfitting of the model. A SoftMax activation function and an Adam Optimizer were used to improve the classification efficiency of the model. This research also adopted a categorical cross-entropy loss function. A summary of the DCNN structure is shown in Table 3 below.

*Table 3. Summary of the Proposed DCNN Layers.*

| Layer (Type) | Output Shape | Param # |
|---|---|---|
| Rescaling (Rescaling) | (None, 48, 48, 1) | 0 |
| sequential (Sequential) | (None, 48, 48, 1) | 0 |
| conv2d_7 (Conv2D) | (None, 46, 46, 32) | 320 |
| conv2d_8 (Conv2D) | (None, 44, 44, 64) | 18496 |
| max_pooling2d_4 (MaxPooling2 | (None, 22, 22, 64) | 0 |
| dropout_3 (Dropout) | (None, 22, 22, 64) | 0 |
| conv2d_9 (Conv2D) | (None, 20, 20, 64) | 36928 |
| conv2d_10 (Conv2D) | (None, 18, 18, 64) | 36928 |
| conv2d_11 (Conv2D) | (None, 16, 16, 128) | 73856 |
| max_pooling2d_5 (MaxPooling2 | (None, 8, 8, 128) | 0 |
| conv2d_12 (Conv2D) | (None, 6, 6, 128) | 147584 |
| conv2d_13 (Conv2D) | (None, 4, 4, 256) | 295168 |
| max_pooling2d_6 (MaxPooling2 | (None, 2, 2, 256) | 0 |
| max_pooling2d_7 (MaxPooling2 | (None, 1, 1, 256) | 0 |
| dropout_4 (Dropout) | (None, 1, 1, 256) | 0 |
| flatten_1 (Flatten) | (None, 256) | 0 |
| dense_2 (Dense) | (None, 1024) | 263168 |
| dropout_5 (Dropout) | (None, 1024) | 0 |
| dense_3 (Dense) | (None, 7) | 7175 |
| Total params: | | 879,623 |
| Trainable params: | | 879,623 |
| Non-trainable params: | | 0 |

Image Data Augmentation: For efficient generalization and optimized model performance, Augmentation was used to expand the size of the training dataset by creating several modified variations of the images using the ImageDataGenerator Class in Keras. Here a target image was converted in an index of 0 to Augmented variations, then looping it through Nine instances of different emotions to see what predictability would look like during the test, as illustrated in Figure 4 below.

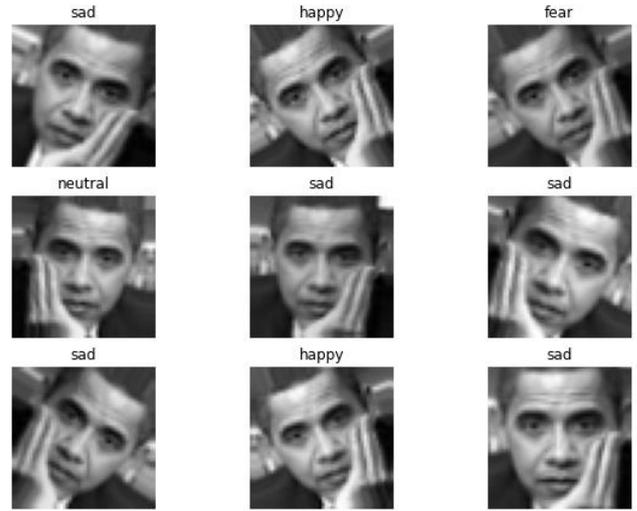

*Figure 4. Result of image Data Augmentation.*

Hyperparameter Tuning: A hyperparameter is a parameter whose value is used to control the training process of a model (neural network). Hyperparameter optimization entails choosing a set of optimal hyperparameters for a learning algorithm to improve the generalization or predictability of a model.

For this research, the random search technique is used wherein random combinations of a range of values for each hyperparameter have been used to find the best possible combination. The effectiveness of the random search technique for hyperparameter optimization has been demonstrated in [24].

*Table 4. Describes all the hyperparameters used by the proposed model and their corresponding values.*

| Hyperparameter | Value | Description |
|---|---|---|
| Batch size | 120 | The number of images (from the training set) to be propagated through the network at a go. |
| Number of epochs | 80 | The number of complete passes through the training set. |
| Optimizer | Adam | Adam Optimization algorithm. |
| Learning rate (lr) | 0.001 | Controls the speed at which weights of the network are updated during training. |
| FC1 neurons | 1024 | Total number of neurons in the first fully connected layer |
| FC2 neurons | 512 | Total number of neurons in the second fully connected layer |
| Dropout | 0.5 | Dropout rate between fully connected layers |
| Convolution kernel size | 3x3 | Size of the kernel in each convolution layer |
| MaxPooling kernel size | 2x2 | Size of the kernel in each MaxPooling layer |
| MaxPooling strides | 2 | Kernel stride in each MaxPooling layer |

## 6. Result Analysis and Discussion

The algorithm and Facial Emotion Recognition model proposed by this research is based on two principal ideas; Firstly, the utilization of high-capacity deep convolutional neural networks for feature extraction and emotion classification, which employs a single classifier for detecting faces from multiple views over both real-time scenarios and digital images or video frames. This research also sought to optimize the computational complexity of the Deep Convolutional Neural Network (DCNN) by modifying the architecture through the addition of layers to improve pattern identification in real-time or digital images. The additional layers apply more convolution filters to the image to detect features of the image. To further enhance the model's predictive efficiency and accuracy, the number of training Epoch was increased to 80.

In emotion recognition, 3 steps, i.e., face detection (See Figure 3), features extraction, and emotion classification using deep learning was employed in the proposed model, which offers higher-end results than the preceding model. In the proposed method, as computation time reduces, validation accuracy increases, with a significant reduction in validation loss. The proposed DCNN model was tested on the FER-2013 dataset (see Table 1), encompassing the seven primary emotions (sad, fear, happiness, angry, neutral, surprised, and disgust).

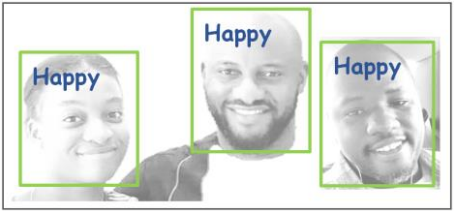

*Figure 5.* Result of Real-time Test Sample associated with the HAPPY Emotion.

Figure 5 above indicates the test sample result associated with happy emotions from the digital test image. The proposed model also predicted the identical emotion with decreased computation time compared to preceding models discussed in the literature review.

*Table 5.* Showing the Evaluation Metrics for the Proposed CNN.

|              | Precision | Recall | F1-Score | Support |
|--------------|-----------|--------|----------|---------|
| angry        | 0.62      | 0.64   | 0.63     | 958     |
| disgust      | 0.73      | 0.34   | 0.45     | 111     |
| fear         | 0.56      | 0.42   | 0.48     | 1024    |
| happy        | 0.91      | 0.91   | 0.91     | 1774    |
| neutral      | 0.64      | 0.71   | 0.67     | 1233    |
| sad          | 0.56      | 0.62   | 0.59     | 1247    |
| surprise     | 0.79      | 0.81   | 0.72     | 831     |
| accuracy     |           |        | 0.70     | 7178    |
| macro avg    | 0.69      | 0.65   | 0.66     | 7178    |
| weighted avg | 0.70      | 0.70   | 0.70     | 7178    |

Table 5 above describes the metrics used in measuring the success of the CNN model developed for this research. Given the necessary modifications made as proposed earlier, it was observed that, on average, the proposed model had a predictive accuracy of 70%. The weighted average of the test dataset is also 70%.

After the training was done, the model was evaluated and the training and validation accuracy (Figure 6) and loss (Figure 7) were computed.

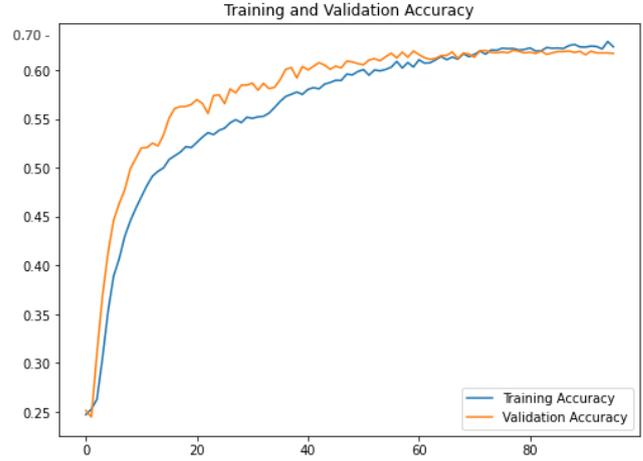

*Figure 6.* Plots of the Training and Validation Accuracy.

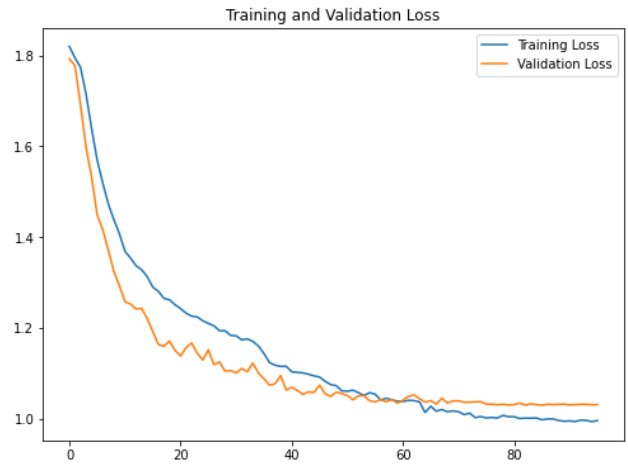

*Figure 7.* Plot of the Training and Validation Loss.

Figure 6 and Figure 7 show the CNN model's learning curve, which plots the model's learning performance over experience or time. After each update, the model was evaluated on the training dataset and a hold-out validation dataset. Learning curves were preferred as a graphic metric for this research because of their wide adoption in machine learning for models that optimize their internal parameters incrementally. From the Learning Curve in Figure 7, it can be observed that the plot of training loss and validation loss declines to the point of stability with a generalization gap of minimal difference. Also, it can be inferred in Figure 6 that the plot of training accuracy and validation accuracy surges with an increase in training sequence and batch size with a minimal generalization gap. Thematically, the model can be a good fit and is proposed to generalize efficiently.

Figure 8 shows the CNN model's confusion matrix on the FER2013 testing set. It was inferred that the model generalizes best in the classification of "happiness" and "surprise" emotions. In contrast, it performs reasonably average when classifying "disgust" and "fear" Emotion. This reduction in classification accuracy in "disgust" and "fear" emotions can be ascribed to the reduced number of training set samples for the two classes. The complacency between

"fear" and "sadness" may be due to the inter-class similarities of the dataset

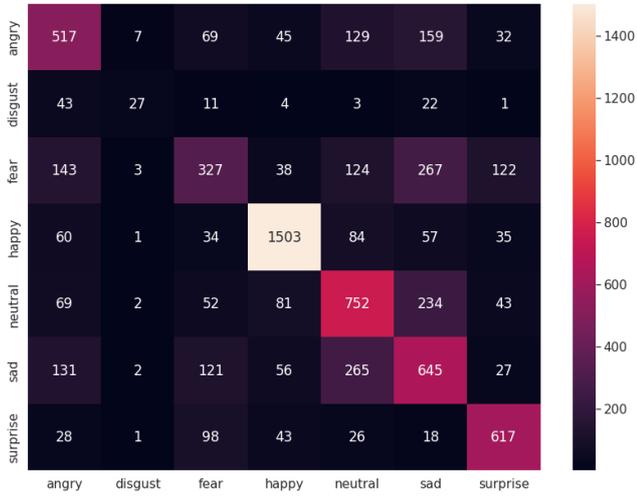

*Figure 8.* Confusion Metrics of the Model on Emotion Prediction.

## 7. Conclusion

This paper proposed a deep CNN model for Real-time Facial Expression Recognition based on seven emotional classes ('neutral', 'happy', 'sad', 'angry', 'surprised', 'fear', and 'disgusted'). The structure of the proposed model has good generality and classification performance. Firstly, a variety of well-classified, high-quality databases were acquired. Then, the face region is detected, cut, and converted into a grayscale image of one channel to remove unnecessary information. Image Data Augmentation which increases the number and variations of training images is applied to solve the problem of overfitting. Hyperparameter tuning was employed to achieve a state-of-the-art classification performance accuracy of 70.04%.

In the proposed hybrid architecture, an optimal structure to reduce execution time and improve the classification performance of real-time and digital images was developed. This was done by adjusting the number of feature maps in the convolutional layer, layers in the Neural Network Model, and numerous training epochs. Cross-validation experiments showed that the proposed Convolutional Neural Network (CNN) Architecture has better classification performance and generality than some state-of-the-arts (SoTA). The Haar Cascade model employed for real-time facial detection showed better classification performance than other implementations. Experimental results confirmed the effectiveness of data preprocessing and augmentation techniques.

Some shortcomings in this research were a deficiency in predicting the 'disgust' and 'angry' emotions due to insufficient training dataset for the two-class categories. Another issue the proposed model faces is the reduced generality of real-time predictions. A major causative factor stems from the postured nature of the training images and environmental conditioning (lightening) of real-time test images. A significant impact of this research in Human-Computer Interaction (HCI) will be the upscaling of software and AI systems to deliver an improved experience to humans in various applications, for instance, in Home robotic systems in the recommendation of mood-based music. This research will enhance Psychological Prognosis by assisting psychologists in detecting probable suicides and emotional traumas, and it can also be employed in the marketing, healthcare, and gaming industry.

## Future Research Proposal

Human facial expression recognition is an important technique; it is beneficial to assess a topic's mood or emotional state underneath observation. The simple concept of a machine recognizing the human emotive state may be used in numerous eventualities. As such, numerous potentials remain untapped in this area. State-of-the-Art real-time algorithms (Faster R-CNN, HOG + Linear SVM, SSDs, YOLO) are encouraged to be employed in future research as well. In addition, advancement to this research may include the utilization of regression in analyzing Bio-signals and physiological multimodal features in determining the levels and intensities of emotion classes discussed herein.

## Abbreviations

DCNN - Deep Convolutional Neural Network
CNN - Convolutional Neural Network
FER - Facial Emotion Recognition
GPU - Graphical Processing Unit
HCI - Human Computer Interaction
ECG - Electrocardiography
FMRI - Functional Magnetic Resonance Imaging
AFEW - Acted Facial Expressions in the Wild
RNN - Recurrent Neural Network
LSTM - Long Short-Term Memory
HMM - Hidden Markov Model
ReLU - Rectified Linear Unit
TFE - Tandem Facial Expression
ANN - Artificial Neural Network
SVM - Support Vector Machine
IRNN - Image Recognition Neural Network
FERET - Facial Recognition Technology
SoTA - State-of-The-Art

## Disclosure

The authors declare that they have no competing interests.

## Author Contributions

All authors made significant contributions to conception and design, data acquisition, analysis and interpretation; participated in drafting the article and critically revising it for important intellectual content; agreed to submit it to the current journal; gave final approval of the version to be published.

## Data Availability Statement

This research paper analyzes the Facial Emotion Recognition (FER-2013) dataset housed by the Kaggle repository [https://www.kaggle.com/datasets/msambare/fer2013].